\pdfoutput=1
\documentclass{gdm-report}

\usepackage{multirow}
\usepackage{colortbl}
\usepackage{xspace}
\usepackage{placeins}
\usepackage{pgfplots}
\usetikzlibrary{arrows.meta,positioning,fit,calc,backgrounds,shapes.geometric}
\pgfplotsset{compat=1.18}

\definecolor{vteBlue}{HTML}{35618D}       
\definecolor{vteTeal}{HTML}{246D73}       
\definecolor{vteOchre}{HTML}{B77C35}      
\definecolor{vteNeutral}{HTML}{5E666E}    
\definecolor{vteBlueLight}{HTML}{F3F6F9}
\definecolor{vteTealLight}{HTML}{EFF6F6}
\definecolor{vteOchreLight}{HTML}{FBF6EE}
\definecolor{vteMasked}{HTML}{E6E9EC}
\definecolor{vteRule}{HTML}{D0D6DC}

\newcommand{\cutcm}{Cut-CrossModal\xspace}
\newcommand{\cutcf}{Cut-CrossFrame\xspace}
\newcommand{\cutboth}{Cut-Both\xspace}
\newcommand{\mftd}{MFTD\xspace}
\newcommand{\vidindex}{Video-Index\xspace}

\newif\ifisICML
\isICMLfalse

\isICMLfalse
\reporttag{Technical Report}
\title{Video MLLMs Are Implicitly Early In-Fusion:\\Architectural Lessons for Native Video Training}
\author{\centering {\color{KeyeInk}\fontsize{11}{13.5}\selectfont Keye-Video Team}}
\headertext{Video MLLMs Are Implicitly Early In-Fusion}
\reportdate{November 26, 2026}
\reportabstract{Most video multimodal large language models (MLLMs) process vision with a standalone encoder, or duplicate modality-specific parameters in every layer. Recent work on static images shows that placing dedicated visual capacity only in the early decoder layers, a design called \emph{early in-fusion}, is sufficient. Video additionally requires binding information across frames, and whether early in-fusion extends to video remains open. In this paper, we examine where this binding occurs in existing video MLLMs. We analyze five open-weights models from the Gemma~4 and Qwen3-VL families, spanning encoder-free, ViT-equipped, dense, and sparse MoE designs. Using training-free prefix attention knockouts and forward-versus-reversed frame drift on Video-Index, we find three consistent patterns. First, cross-frame attention matters only in the lower half of every model: blocking it beyond half depth shifts decision preservation by at most 3.1 points. Second, in encoder-free Gemma~4~12B, the first eighth of layers encodes each frame independently of other frames and of the prompt, retaining 91.2\% of baseline decisions under joint isolation. Under the same isolation, however, Qwen3-VL loses 16--23 points more than under either single-axis knockout, for reasons not yet determined. Third, in hybrid sliding-window decoders, temporal drift peaks at global attention layers. Together, these measurements inform three design choices for native video MLLMs: how deep dedicated visual capacity should extend, when joint visual isolation is safe, and where global attention layers belong.
}

\begin{document}
\maketitle

\section{Introduction}
\label{sec:intro}

Most video multimodal large language models (MLLMs) attach a separately pretrained vision encoder~\citep{dosovitskiy2021vit} to a language model~\citep{bai2025qwen3vl,team2026kwai,yu2026minicpm,wang2024cogvlm}, or duplicate modality-specific pathways in every transformer layer~\citep{vaswani2017attention,luo2025monointernvl,diao2025evev2,diao2026pixels,liang2024mot}. Recent studies on static images question both designs. \citet{harwath2026implicit} show that the early layers of encoder-free models form context-invariant implicit encoders. Under compute-matched pretraining, \citet{brown2026unbundling} show that dedicated visual capacity placed only in the early decoder layers, termed \emph{early in-fusion}, matches a separate vision tower and outperforms full-depth modality separation.

Video poses a question that static images do not. Even when frames enter the decoder with timestamps~\citep{gemma2026gemma4,bai2025qwen3vl}, the model must perceive each frame and relate frames across space and time~\citep{bertasius2021timesformer,arnab2021vivit}. We refer to this cross-frame integration as \emph{inter-frame temporal binding}. If early in-fusion is to anchor native video training, this binding must also complete in the lower layers. We therefore ask:
\begin{quote}
\emph{Do existing video MLLMs confine frame encoding and inter-frame temporal binding to their lower layers, and which architectural factors matter for native early in-fusion video training?}
\end{quote}

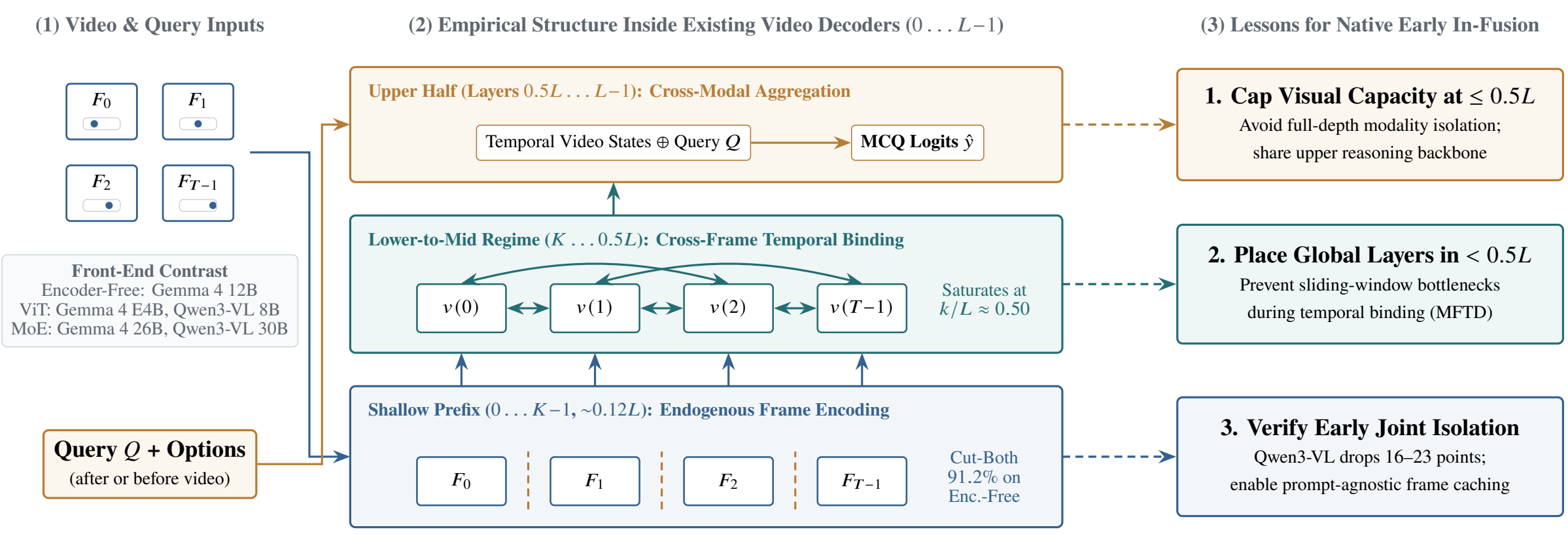
\begin{figure*}[t]
\centering
\resizebox{\textwidth}{!}{%
\begin{tikzpicture}[
    font=\small,
    >=Stealth,
    frame/.style={draw=vteBlue, line width=0.7pt, rounded corners=1.5pt, fill=white, minimum width=9.5mm, minimum height=7.5mm, inner sep=1pt, align=center},
    qbox/.style={draw=vteOchre, line width=0.8pt, rounded corners=2pt, fill=vteOchreLight, minimum width=24mm, minimum height=9mm, inner sep=4pt, align=center},
    bandEarly/.style={draw=vteBlue, line width=0.8pt, rounded corners=2pt, fill=vteBlueLight},
    bandMid/.style={draw=vteTeal, line width=0.8pt, rounded corners=2pt, fill=vteTealLight},
    bandUpper/.style={draw=vteOchre, line width=0.8pt, rounded corners=2pt, fill=vteOchreLight},
    lane/.style={draw=vteBlue, line width=0.65pt, rounded corners=1.5pt, fill=white, minimum width=12mm, minimum height=6.5mm, inner sep=1pt, align=center, font=\scriptsize},
    guidebox/.style={draw=vteTeal, line width=0.8pt, rounded corners=2pt, fill=white, align=center, inner sep=4.5pt},
    arr/.style={->, line width=0.75pt, draw=vteNeutral},
    varr/.style={->, line width=0.8pt, draw=vteBlue},
    tarr/.style={<->, line width=0.8pt, draw=vteTeal},
    qarr/.style={->, line width=0.8pt, draw=vteOchre},
]

\node[font=\footnotesize\bfseries, text=vteNeutral] at (0.0, 4.25) {(1) Video \& Query Inputs};

\node[frame] (f0) at (-0.65, 3.10) {\scriptsize $F_0$\\[-2pt]\tikz{\fill[vteBlue] (0,0) circle (1.4pt); \draw[vteRule] (-0.15,-0.08) rectangle (0.35,0.08);}};
\node[frame] (f1) at (0.65,  3.10) {\scriptsize $F_1$\\[-2pt]\tikz{\fill[vteBlue] (0.10,0) circle (1.4pt); \draw[vteRule] (-0.15,-0.08) rectangle (0.35,0.08);}};
\node[frame] (f2) at (-0.65, 2.00) {\scriptsize $F_2$\\[-2pt]\tikz{\fill[vteBlue] (0.20,0) circle (1.4pt); \draw[vteRule] (-0.15,-0.08) rectangle (0.35,0.08);}};
\node[frame] (f3) at (0.65,  2.00) {\scriptsize $F_{T-1}$\\[-2pt]\tikz{\fill[vteBlue] (0.30,0) circle (1.4pt); \draw[vteRule] (-0.15,-0.08) rectangle (0.35,0.08);}};

\node[draw=vteRule, rounded corners=2pt, fill=vteBlueLight!60, align=center, font=\scriptsize, inner sep=3.5pt, text=vteNeutral] (front) at (0.0, 0.55) {%
    \textbf{Front-End Contrast}\\[-1pt]
    Encoder-Free: Gemma~4~12B\\[-1pt]
    ViT: Gemma~4~E4B, Qwen3-VL~8B\\[-1pt]
    MoE: Gemma~4~26B, Qwen3-VL~30B};

\node[qbox] (query) at (0.0, -1.65) {\textbf{Query $Q$ + Options}\\[-1pt]{\scriptsize (after or before video)}};

\node[font=\footnotesize\bfseries, text=vteNeutral] at (7.5, 4.25) {(2) Empirical Structure Inside Existing Video Decoders ($0 \dots L{-}1$)};

\draw[bandUpper] (2.7, 2.15) rectangle (12.3, 3.70);
\coordinate (upper_w) at (2.7, 2.925);
\coordinate (upper_e) at (12.3, 2.925);
\node[anchor=north west, font=\scriptsize\bfseries, text=vteOchre] at (2.82, 3.60) {Upper Half (Layers $0.5L \dots L{-}1$): Cross-Modal Aggregation};
\node[draw=vteOchre, fill=white, rounded corners=1.5pt, font=\scriptsize, inner sep=3.5pt] (fuse) at (6.25, 2.68) {Temporal Video States $\oplus$ Query $Q$};
\node[draw=vteOchre, fill=white, rounded corners=1.5pt, font=\scriptsize\bfseries, inner sep=3.5pt] (ans) at (10.35, 2.68) {MCQ Logits $\hat{y}$};
\draw[qarr] (fuse.east) -- (ans.west);

\draw[bandMid] (2.7, -0.15) rectangle (12.3, 1.70);
\coordinate (mid_e) at (12.3, 0.775);
\node[anchor=north west, font=\scriptsize\bfseries, text=vteTeal] at (2.82, 1.60) {Lower-to-Mid Regime ($K \dots 0.5L$): Cross-Frame Temporal Binding};
\node[lane, draw=vteTeal] (m0) at (4.2, 0.45) {$v(0)$};
\node[lane, draw=vteTeal] (m1) at (6.0, 0.45) {$v(1)$};
\node[lane, draw=vteTeal] (m2) at (7.8, 0.45) {$v(2)$};
\node[lane, draw=vteTeal] (m3) at (9.6, 0.45) {$v(T{-}1)$};
\draw[tarr] (m0.east) -- (m1.west);
\draw[tarr] (m1.east) -- (m2.west);
\draw[tarr] (m2.east) -- (m3.west);
\draw[tarr] (m0.north) to[bend left=14] (m2.north);
\draw[tarr] (m1.north) to[bend left=14] (m3.north);
\node[font=\scriptsize, text=vteTeal, align=center] at (11.25, 0.55) {Saturates at\\[-1pt]$k/L \approx 0.50$};

\draw[bandEarly] (2.7, -2.50) rectangle (12.3, -0.60);
\coordinate (early_w) at (2.7, -1.55);
\coordinate (early_e) at (12.3, -1.55);
\node[anchor=north west, font=\scriptsize\bfseries, text=vteBlue] at (2.82, -0.70) {Shallow Prefix ($0 \dots K{-}1$, ${\sim}0.12L$): Endogenous Frame Encoding};
\node[lane] (e0) at (4.2, -1.88) {$F_0$};
\node[lane] (e1) at (6.0, -1.88) {$F_1$};
\node[lane] (e2) at (7.8, -1.88) {$F_2$};
\node[lane] (e3) at (9.6, -1.88) {$F_{T-1}$};

\foreach \x in {5.1, 6.9, 8.7} {
    \draw[densely dashed, line width=0.8pt, draw=vteOchre] (\x, -2.26) -- (\x, -1.48);
}
\node[font=\scriptsize, text=vteBlue, align=center] at (11.25, -1.82) {\cutboth{}\\[-1pt]$91.2\%$ on\\[-1pt]Enc.-Free};

\foreach \x in {4.2, 6.0, 7.8, 9.6} {
    \draw[varr] (\x, -0.60) -- (\x, -0.15);
}
\draw[->, line width=0.85pt, draw=vteTeal] (6.25, 1.70) -- (6.25, 2.15);

\draw[varr] (1.35, 2.55) -- (2.15, 2.55) |- (early_w);
\draw[qarr] (query.east) -- (2.32, -1.65) |- (upper_w);

\node[font=\footnotesize\bfseries, text=vteNeutral] at (16.45, 4.25) {(3) Lessons for Native Early In-Fusion};

\node[guidebox, draw=vteOchre, fill=vteOchreLight, minimum width=52mm, minimum height=15mm] (g_upper) at (16.45, 2.925) {%
    \textbf{1. Cap Visual Capacity at $\le 0.5L$}\\[-1pt]
    {\scriptsize Avoid full-depth modality isolation;}\\[-1pt]
    {\scriptsize share upper reasoning backbone}};

\node[guidebox, fill=vteTealLight, minimum width=52mm, minimum height=16mm] (g_mid) at (16.45, 0.775) {%
    \textbf{2. Place Global Layers in $< 0.5L$}\\[-1pt]
    {\scriptsize Prevent sliding-window bottlenecks}\\[-1pt]
    {\scriptsize during temporal binding (\mftd)}};

\node[guidebox, draw=vteBlue, fill=vteBlueLight, minimum width=52mm, minimum height=16mm] (g_early) at (16.45, -1.55) {%
    \textbf{3. Verify Early Joint Isolation}\\[-1pt]
    {\scriptsize Qwen3-VL drops $16\text{--}23$ points;}\\[-1pt]
    {\scriptsize enable prompt-agnostic frame caching}};

\draw[->, line width=0.85pt, densely dashed, draw=vteOchre] (upper_e) -- (g_upper.west);
\draw[->, line width=0.85pt, densely dashed, draw=vteTeal]  (mid_e)   -- (g_mid.west);
\draw[->, line width=0.85pt, densely dashed, draw=vteBlue]  (early_e) -- (g_early.west);

\end{tikzpicture}%
}
\caption{\textbf{From mechanistic diagnosis inside existing video MLLMs to design lessons for native early in-fusion training.} \textbf{(1)} Timestamped frames $F_0 \dots F_{T-1}$ and query $Q$ enter five models across encoder-free, ViT, and sparse MoE designs. \textbf{(2)} Prefix knockouts and \mftd probing show that existing decoders already confine frame-level encoding ($0 \dots K{-}1$, ${\sim}0.12L$) and cross-frame temporal binding ($K \dots 0.5L$) to the lower half. \textbf{(3)} The measurements give three guidelines for training future native early in-fusion video models.}
\label{fig:overview}
\end{figure*}

To answer this question, we separate two kinds of attention inside the decoder: \emph{cross-modal} attention between visual and textual tokens, and \emph{cross-frame} attention between visual tokens of different frames. As illustrated in \Cref{fig:overview}, we block each pathway in a prefix of layers and measure how predictions change. If frames are encoded independently in the shallowest layers and cross-frame binding finishes by half depth, dedicated visual capacity need only cover the lower half of the network.

Such measurements require a benchmark that cannot be solved from a single frame or from the question alone. We use a 160-item stratified subset of \vidindex~\citep{song2026videoindex}, filtered from 115 benchmarks by an attack pyramid using Qwen3-VL~8B to remove shortcut-solvable items. Baseline zero-CoT accuracy on this subset is low, ranging from $7.50\%$ to $20.00\%$. Alongside accuracy, we therefore report two measures: the \emph{Decision Preservation Rate} $\mathrm{DP}_m(k)$, which tracks how often predictions stay unchanged under intervention, and \emph{Matched-Frame Temporal Drift}, or \mftd, which measures how frame representations shift between forward and reversed playback.

We study five open-weights models. ViT-equipped Gemma~4~E4B-it and encoder-free Gemma~4~12B-it form an intra-family encoder contrast. Gemma~4~26B-A4B-it and Qwen3-VL~30B-A3B represent sparse MoE designs~\citep{shazeer2017moe}, and Qwen3-VL~8B-Instruct provides a full-attention dense baseline. Our findings are as follows.
\begin{enumerate}
    \item \textbf{Cross-frame attention matters only in the lower half.} In all five models, \cutcf decision preservation drops over the lower half and stays flat beyond $k/L \approx 0.50$. On Gemma~4~26B-A4B, \emph{Temporal} accuracy falls from $20.0\%$ to $5.0\%$ before the midpoint.
    \item \textbf{Joint isolation of early layers depends on the model.} Under \cutboth masking across its first eighth of layers, $k/L{=}0.125$, encoder-free Gemma~4~12B keeps $91.2\%$ of its decisions and $95.8\%$ of its accuracy. Under the same masking, ViT-equipped Qwen3-VL~8B drops to $60.0\%$ at $k/L{=}0.11$ and $58.8\%$ at $k/L{=}0.25$, below the $71.9\%$ obtained when vision is blocked in all 36 layers. Its cause has not been determined.
    \item \textbf{Attention topology determines where temporal drift appears.} In full-attention Qwen3-VL, \mftd forms a single peak near mid-depth, at $\ell/L{=}0.47$ on 8B and $0.58$ on 30B-A3B. In Gemma~4, which interleaves five local layers with each global layer, \mftd jumps at global layers, increasing tenfold at layer $12$ on 12B.
    \item \textbf{Benchmark and prompt audit.} On $84$ of the $160$ items, at least four of the five models pick the same option, yet only $8.3\%$ of those consensus choices are correct. A prompt hint listing only A--D lowers accuracy on the $21.9\%$ of items whose answer lies in E--H. We release fixes for both the prompt and the probe evaluation.
\end{enumerate}

\section{Related Work}
\label{sec:related}

\paragraph{Vision encoders, encoder-free MLLMs, and early in-fusion.}
Most vision-language models connect a separately pretrained vision encoder~\citep{dosovitskiy2021vit,radford2021clip} to a language model~\citep{liu2023llava,wang2024cogvlm}. This modular recipe extends widely to video via spatiotemporal resamplers and sparse long-context attention~\citep{bai2025qwen3vl,yu2026minicpm,team2026kwai}. Monolithic early-fusion and encoder-free models instead project visual tokens directly into one transformer, simplifying the pipeline and scaling natively from images to video~\citep{chameleon2024,luo2025monointernvl,diao2025evev2,diao2026pixels}. \citet{chen2026howfar} study scaling laws for encoder-free pretraining and find that the advantage of a pretrained encoder narrows near $10^{22}$ FLOPs, as early decoder layers specialize for vision. \citet{brown2026unbundling} separate visual pretraining, dedicated vision capacity, and placement under compute-matched training. They show that visual pretraining is optional, whereas dedicated vision capacity in early layers, or \emph{in-fusion}, is necessary for fine-grained perception such as OCR and charts. Mixture-of-Transformers (MoT)~\citep{liang2024mot} and unified understanding--generation models~\citep{deng2025bagel,xie2025showo2} likewise separate modality pathways while sharing self-attention. Unlike those image pretraining studies, we analyze trained video models at inference time.

\paragraph{Implicit encoders and mechanistic analysis of MLLMs.}
Probing and causal interventions in language models show that transformer layers specialize by stage~\citep{tenney2019bert}. Causal tracing and attention-edge knockouts further localize information flow inside autoregressive decoders~\citep{meng2022locating,geva2023dissecting}. Closest to our work, \citet{harwath2026implicit} apply layer-wise probing and causal interventions to encoder-free image and audio MLLMs, including Gemma~4~12B. They find that early layers act as context-invariant implicit encoders whose representations match standalone encoders without cross-modal context. We extend this inquiry in two directions. First, we move from single images to multi-frame video, where inter-frame attention adds a temporal axis paralleling factorized space--time video transformers~\citep{bertasius2021timesformer,arnab2021vivit}. Second, we compare ViT-equipped and encoder-free Gemma~4 variants~\citep{gemma2026gemma4} alongside 3D-RoPE Qwen3-VL~\citep{bai2025qwen3vl}.

\paragraph{Shortcut-resistant video evaluation and prefill reuse.}
Many video benchmarks exhibit strong single-frame or question-only shortcuts~\citep{lei2023singleframe}, which confound mechanistic analysis even on broad MLLM evaluations~\citep{fu2025videomme}. \vidindex~\citep{song2026videoindex} audits 115 video benchmarks with an attack pyramid and retains 840 verified items in four capability groups. Evaluating knockouts on \vidindex lowers the risk that preserved accuracy stems from such shortcuts. On the systems side, prefix KV reuse is central to modern LLM serving~\citep{kwon2023vllm,zheng2024sglang}. If early layers are invariant to other frames and to the prompt, their states can be cached and reused across queries in multi-query video serving and real-time full-duplex interaction, where perceptual streams run alongside evolving conversational context~\citep{cui2026minicpm,zhang2026liberating}. Such reuse avoids visual recomputation whenever preceding text changes.

\section{Method}
\label{sec:method}

Our goal is to locate, inside an $L$-layer multimodal decoder, where frame encoding ends and inter-frame temporal binding begins. We do so without training, by masking attention in a prefix of layers and by comparing representations under reversed frame order.

\subsection{Inputs and Metrics}
\label{sec:method_input}

Given a video $V$ and a multiple-choice question $Q$ with options $\mathcal{O}$, we uniformly sample $T$ frames $[F_0, F_1, \dots, F_{T-1}]$ with increasing timestamps $[\tau_0, \tau_1, \dots, \tau_{T-1}]$. Each frame is preceded by a timestamp marker such as $[\tau_t\text{s}]$, and the model's native processor maps the input to a sequence of $S$ tokens. Let $\mathcal{V}_t = [s_t, e_t) \subset \{0, \dots, S{-}1\}$ be the visual token span of frame $F_t$, including begin/end-of-image markers when present. Let $\mathcal{V} = \bigcup_{t=0}^{T-1} \mathcal{V}_t$ denote all visual tokens and $\mathcal{T} = \{0, \dots, S{-}1\} \setminus \mathcal{V}$ denote all textual tokens.

To avoid free-form decoding variation, we read predictions from next-token logits at the final prompt position:
\begin{equation}
    \hat{y}_n^{(m)}(k) = \arg\max_{c \in \mathcal{O}_n} \operatorname{logsumexp}_{u \in \mathcal{U}(c)} z_{n, u}^{(m)}(k),
\end{equation}
where $z_n^{(m)}(k)$ are the final-position logits for item $n$ under intervention mode $m$ at depth $k$, and $\mathcal{U}(c)$ collects single-token variants of option letter $c$, such as ``A'' and `` A''. Let $\hat{y}_n(0)$ be the unmodified baseline prediction at $k{=}0$. On a shortcut-filtered benchmark, baseline accuracy is low and accuracy differences rest on few items. We therefore report accuracy $\mathrm{Acc}_m(k)$ together with the \emph{Decision Preservation Rate}:
\begin{equation}
    \mathrm{DP}_m(k) = \frac{1}{N} \sum_{n=1}^{N} \mathbb{I}\!\left[\hat{y}_n^{(m)}(k) = \hat{y}_n(0)\right].
    \label{eq:dp}
\end{equation}
$\mathrm{DP}_m(k)$ uses all $N$ items, whether or not the baseline prediction is correct, and measures how often masking layers $0 \dots k{-}1$ leaves the prediction unchanged.

\begin{figure*}[t]
\centering
\includegraphics[width=\textwidth]{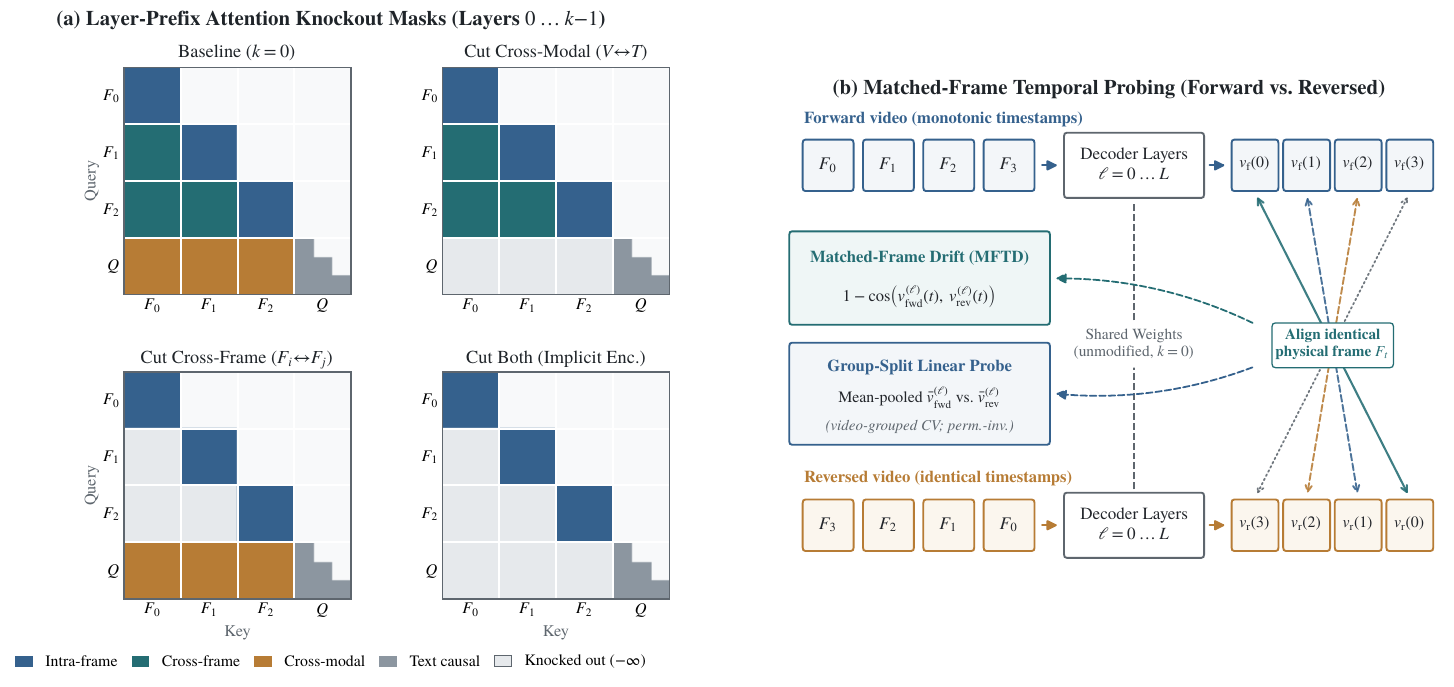}
\caption{\textbf{Diagnostic protocol: structured prefix attention knockouts and matched-frame temporal probing.} \textbf{(a)} The four layer-prefix attention patterns applied across layers $0 \dots k{-}1$ while preserving native causal, sliding-window, and intra-frame bidirectional constraints: \emph{Baseline} ($k{=}0$), \cutcm (blocking $V \leftrightarrow T$), \cutcf (blocking $F_i \leftrightarrow F_j$ for $i \neq j$), and \cutboth (each frame encoded in isolation). Diagonal self-attention is always preserved. \textbf{(b)} Forward $[F_0, \dots, F_3]$ and reversed $[F_3, \dots, F_0]$ passes share monotonic timestamps and unmodified model weights; aligning each physical frame $F_t$ across the two passes yields the Matched-Frame Temporal Drift (\mftd, \Cref{eq:mftd}), while mean-pooling across frames yields a permutation-invariant input for 5-fold linear temporal-direction probing.}
\label{fig:protocol}
\end{figure*}

\subsection{Prefix Attention Knockouts}
\label{sec:method_knockout}

Video MLLMs use layer-specific native attention masks $M_{\text{native}}^{(\ell)} \in \{0, -\infty\}^{S \times S}$ for $\ell \in \{0, \dots, L{-}1\}$. These masks may combine causal structure, alternating local sliding-window and global layers~\citep{beltagy2020longformer,gemma2026gemma4}, and bidirectional attention within each frame. Following attention-edge knockout analysis~\citep{geva2023dissecting}, we keep $M_{\text{native}}^{(\ell)}$ and add a binary disallow mask $D^{(m)} \in \{0, 1\}^{S \times S}$ in the first $k$ layers:
\begin{equation}
    \tilde{M}^{(\ell)}_{i,j} =
    \begin{cases}
        -\infty, & \text{if } \ell < k \text{ and } D^{(m)}_{i,j} = 1,\\
        M_{\text{native}, i, j}^{(\ell)}, & \text{otherwise}.
    \end{cases}
\end{equation}
We set $D^{(m)}_{i,i} = 0$ for self-attention so softmax stays well defined. Layers $\ell \ge k$ are unchanged.

\Cref{fig:protocol}a illustrates the three knockout modes:
\begin{itemize}
    \item \textbf{\cutcm:} $D_{i,j} = 1$ if $(i \in \mathcal{V}, j \in \mathcal{T})$ or $(i \in \mathcal{T}, j \in \mathcal{V})$. In layers $0 \dots k{-}1$, visual and textual tokens do not attend to each other, while attention between frames is kept.
    \item \textbf{\cutcf:} $D_{i,j} = 1$ if $i \in \mathcal{V}_t$, $j \in \mathcal{V}_{t'}$, and $t \neq t'$. In layers $0 \dots k{-}1$, each frame attends only to itself and to text tokens.
    \item \textbf{\cutboth:} $D^{\text{both}} = D^{\text{cm}} \lor D^{\text{cf}}$. In layers $0 \dots k{-}1$, each frame is encoded independently of other frames and all text. Both paths start at layer $k$, so preserved predictions mark an implicit frame encoder.
\end{itemize}

Because models differ in depth $L$, we report normalized depth $k/L \in [0, 1]$. We define the \emph{isolation depth} $K_{\tau}^{(m)}$ as the largest swept $k$ satisfying $\mathrm{DP}_m(k') \ge \tau$ for all $k' \le k$, with $\tau = 85\%$. To compare joint masking against single-axis masking, we define the \emph{super-additive drop} at the shallowest cutoff $k_1 \approx 0.12L$:
\begin{equation}
    \Delta_{\text{sink}}(k_1) = \min\!\big(\mathrm{DP}_{\text{cf}}(k_1),\, \mathrm{DP}_{\text{cm}}(k_1)\big) - \mathrm{DP}_{\text{both}}(k_1).
    \label{eq:dsink}
\end{equation}
A positive value indicates that joint masking alters more decisions than either single-axis mask. In every run, we assert at $k = L$ that the hook fires on exactly $L$ layers and changes the logits relative to $k = 0$.

\subsection{Matched-Frame Temporal Drift}
\label{sec:method_probe}

Knockouts measure whether an attention path is \emph{necessary}. To locate where temporal context enters the representation, \Cref{fig:protocol}b compares each frame across two unmodified forward passes at $k = 0$: one in forward order $[F_0, \dots, F_{T-1}]$ and one in reversed order $[F_{T-1}, \dots, F_0]$, keeping timestamps increasing.

Let $v_{\text{fwd}}^{(\ell)}(t) \in \mathbb{R}^d$ be the mean-pooled state of $\mathcal{V}_t$ at layer $\ell$ in the forward pass, and $v_{\text{rev}}^{(\ell)}(t)$ that of the \emph{same frame} $F_t$ at slot $T{-}1{-}t$ in the reversed pass. The \emph{Matched-Frame Temporal Drift}, \mftd, at layer $\ell \in \{0, \dots, L\}$ is:
\ifisICML
\begin{equation}
\begin{aligned}
    \mathrm{MFTD}(\ell) &= \frac{1}{N T} \sum_{n=1}^{N} \sum_{t=0}^{T-1}
    \Bigg(1 - \\
    &\qquad \frac{\langle v_{\text{fwd}, n}^{(\ell)}(t),\, v_{\text{rev}, n}^{(\ell)}(t) \rangle}
    {\|v_{\text{fwd}, n}^{(\ell)}(t)\|_2 \, \|v_{\text{rev}, n}^{(\ell)}(t)\|_2}\Bigg).
\end{aligned}
    \label{eq:mftd}
\end{equation}
\else
\begin{equation}
    \mathrm{MFTD}(\ell) = \frac{1}{N T} \sum_{n=1}^{N} \sum_{t=0}^{T-1}
    \left(1 - \frac{\langle v_{\text{fwd}, n}^{(\ell)}(t),\, v_{\text{rev}, n}^{(\ell)}(t) \rangle}
    {\|v_{\text{fwd}, n}^{(\ell)}(t)\|_2 \, \|v_{\text{rev}, n}^{(\ell)}(t)\|_2}\right).
    \label{eq:mftd}
\end{equation}
\fi
If layers $0 \dots \ell{-}1$ encode each frame independently of its neighbors and its position, $\mathrm{MFTD}(\ell) \approx 0$. Rising drift marks the layers where temporal context alters the frame representation.

\paragraph{Linear probing.}
A linear classifier on frame-averaged features $\bar{v}^{(\ell)} = \frac{1}{T}\sum_t v^{(\ell)}(t)$ can also test whether a layer encodes temporal direction, provided cross-validation folds are grouped by video. The forward and reversed passes of a video share identical layer-0 features, $\bar{v}_{\text{fwd}}^{(0)} = \bar{v}_{\text{rev}}^{(0)}$, yet carry opposite labels. Ungrouped cross-validation therefore drives validation accuracy far below chance, reaching $7.5\%$ at $\ell{=}0$ in our runs. As detailed in \Cref{app:impl}, we omit the ungrouped probe and report \mftd, $\mathrm{DP}_m(k)$, and the reversal drop $\Delta_{\text{rev}} = \mathrm{Acc}_{\text{fwd}} - \mathrm{Acc}_{\text{rev}}$. Our released pipeline includes a corrected probe using \texttt{GroupKFold}.

\subsection{Implications for Training and Caching}
\label{sec:method_cache}

The three knockout modes bear on two practical questions. For \textbf{native early in-fusion pretraining}, the depth where $\mathrm{DP}_{\text{cf}}(k)$ stops changing indicates how deep dedicated visual-to-visual parameters should extend, and $\Delta_{\text{sink}}(k_1)$ tests whether isolating early visual tokens from both frames and text is safe. For \textbf{inference serving}, if a model has a \cutboth isolation depth $K = K_\tau^{\text{both}}$, its layer-$K$ frame states $\{H_t^{(K)}\}$ depend neither on other frames nor on text and can be cached across queries $Q_1, Q_2, \dots$. If only \cutcf holds up to $K_\tau^{\text{cf}}$, frames can still be encoded in parallel up to that depth for streaming prefill.

\section{Experiments}
\label{sec:experiments}

\subsection{Experimental Setup}
We evaluate on a fixed stratified subset of $N = 160$ items from \vidindex~\citep{song2026videoindex}, with $40$ items from each capability group: Perception, Reasoning \& Knowledge, Spatial \& Physical, and Temporal. From each video we uniformly sample $T = 16$ frames, resize them to a maximum side of $384$ pixels, prefix each with its timestamp $[\tau_t\text{s}]$, and place the video before the question. All models run in bfloat16 without quantization using SDPA attention. For Gemma~4~E4B, whose audio tower requires eager attention, we select SDPA or eager per submodule. At runtime we assert that the visual token spans match the $16$ frames, that masking changes the logits, and that the hook fires on exactly $L$ layers at $k = L$. Reversal experiments in \Cref{eq:mftd} use the first $N = 120$ items.

\begin{table*}[t]
\centering
\caption{\textbf{Cross-model mechanistic summary on \vidindex ($N{=}160$ stratified pilot, $T{=}16$ frames, BF16; reversal probing on $N{=}120$).} \textbf{Base} is unmodified MCQ accuracy ($k{=}0$); \textbf{Both@$L$} is full-depth \cutboth blind accuracy. Decision preservation $\mathrm{DP}_m(k)$ reports the percentage of items whose predicted option matches the $k{=}0$ prediction under \cutcf (shown at $k_1 / k_{0.5} / L$, where $k_1 \approx 0.12L$ and $k_{0.5} \approx 0.50L$) and \cutboth (shown at $k_1 / L$). $\Delta_{\text{rev}} = \mathrm{Acc}_{\text{fwd}} - \mathrm{Acc}_{\text{rev}}$ is the behavioral accuracy drop upon frame reversal, and \textbf{Int.\ Peak \mftd} is the largest interior ($\ell < L$) drift with its depth $\ell/L$.}
\label{tab:main_results}
\footnotesize
\setlength{\tabcolsep}{4.5pt}
\begin{tabular}{@{}llccccccc@{}}
\toprule
\textbf{Model} & \textbf{Front-End} & $L$ & \textbf{Base} & \textbf{Both@$L$} & \textbf{$\mathrm{DP}_{\text{cf}}$ (\%)} & \textbf{$\mathrm{DP}_{\text{both}}$ (\%)} & $\Delta_{\text{rev}}$ & \textbf{Int.\ Peak \mftd} \\
 & & & (\%) & (\%) & ($k_1 / k_{0.5} / L$) & ($k_1 / L$) & (\%) & ($\ell/L$) \\
\midrule
Gemma~4~E4B-it & $\sim$150M ViT + PLE & 42 & 7.50 & 6.88 & 95.0 / 76.2 / 76.9 & 85.0 / 63.1 & $-3.33$ & 0.0574 (0.57) \\
Gemma~4~12B-it & Encoder-Free & 48 & 15.00 & 10.62 & 90.6 / 80.6 / 80.0 & 91.2 / 53.1 & $+4.17$ & 0.0297 (0.98) \\
Gemma~4~26B-A4B-it & MoE (4B active) & 30 & 20.00 & 9.38 & 81.9 / 68.1 / 68.1 & 80.0 / 49.4 & $+5.83$ & 0.0587 (0.63) \\
Qwen3-VL~8B & ViT + 3D-MRoPE & 36 & 10.00 & 3.12 & 90.6 / 76.9 / 78.1 & 60.0 / 71.9 & $+4.17$ & 0.0787 (0.47) \\
Qwen3-VL~30B-A3B & MoE (3B) + ViT & 48 & 10.62 & 6.88 & 89.4 / 77.5 / 74.4 & 70.0 / 57.5 & $+2.50$ & 0.0315 (0.58) \\
\bottomrule
\end{tabular}
\end{table*}

\begin{figure*}[t]
\centering
\includegraphics[width=\textwidth]{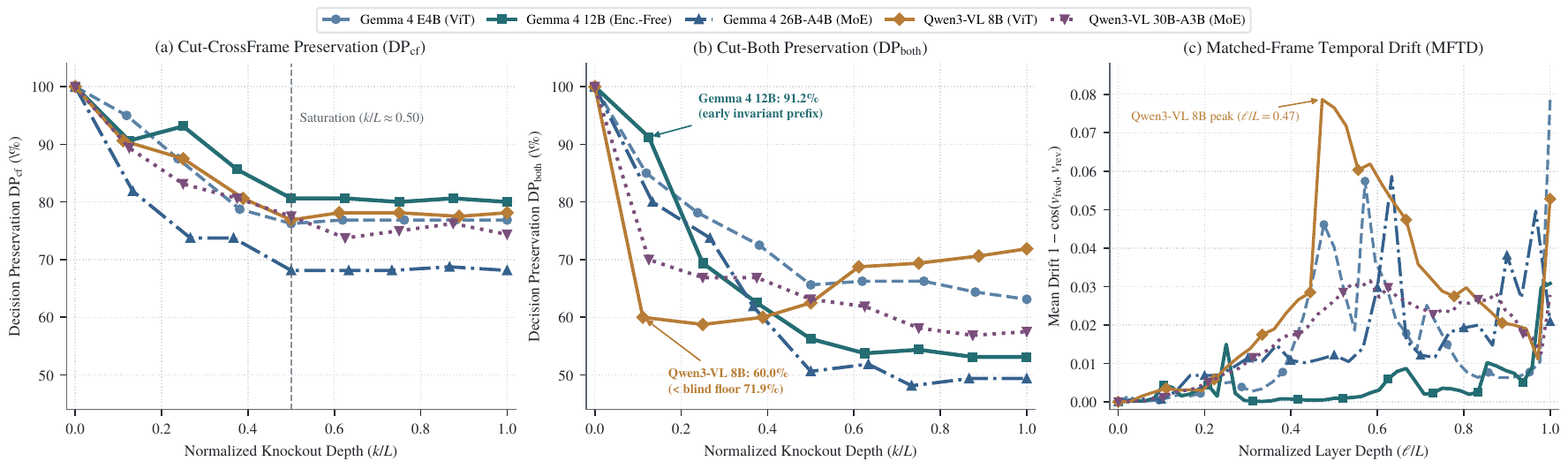}
\caption{\textbf{Cross-model mechanistic profiles on \vidindex ($N{=}160$, $T{=}16$ frames, BF16).} \textbf{(a)} \cutcf decision preservation $\mathrm{DP}_{\text{cf}}(k)$ stops changing at $k/L \approx 0.50$ in all five models (Gemma~4 E4B/12B/26B-A4B and Qwen3-VL 8B/30B-A3B); blocking cross-frame visual attention in the upper half changes few decisions. \textbf{(b)} \cutboth decision preservation separates encoder-free Gemma~4~12B—which retains $91.2\%$ of decisions at $k/L{=}0.125$ ($95.8\%$ accuracy retention)—from ViT-equipped Qwen3-VL~8B, which shows a super-additive drop to $60.0\%$, below its full-depth blind value ($71.9\%$). \textbf{(c)} Matched-Frame Temporal Drift (\mftd) has a single mid-depth peak in Qwen3-VL ($0.0787$ at $\ell/L{=}0.47$ for 8B; $0.0315$ at $\ell/L{=}0.58$ for 30B-A3B) and isolated peaks at global attention layers in hybrid Gemma~4 models.}
\label{fig:main_results}
\end{figure*}

\subsection{Cross-Frame Attention Saturates by Half Depth}
\label{sec:exp_cf}

\Cref{tab:main_results} and \Cref{fig:main_results}a show the same pattern across all five models, despite differences in size from $4\text{B}$ to $30\text{B}$, dense versus MoE routing, and vision front-ends. Under \cutcf, $\mathrm{DP}_{\text{cf}}(k)$ starts near $90\%$ at the shallowest cutoff $k_1 \approx 0.12L$, ranging from $81.9\%$ on Gemma~4~26B-A4B to $95.0\%$ on Gemma~4~E4B. It then decreases monotonically across the lower half and flattens near $k/L = 0.50$.

Between $k/L = 0.50$ and $1.00$, $\mathrm{DP}_{\text{cf}}(k)$ shifts by at most $3.1$ points in any model, as listed in \Cref{tab:main_results}. For example, Gemma~4~26B-A4B stays flat at $68.1\%$, Gemma~4~12B moves only from $80.6\%$ to $80.0\%$, and Qwen3-VL~8B shifts from $76.9\%$ to $78.1\%$. \cutcm follows a similar curve in the lower half before settling at the all-layer blind floor between $53.1\%$ and $71.9\%$. Cross-frame attention thus alters predictions in layers $0 \dots 0.5L$, whereas blocking it above half depth changes few additional decisions.

\begin{figure*}[t]
\centering
\includegraphics[width=\textwidth]{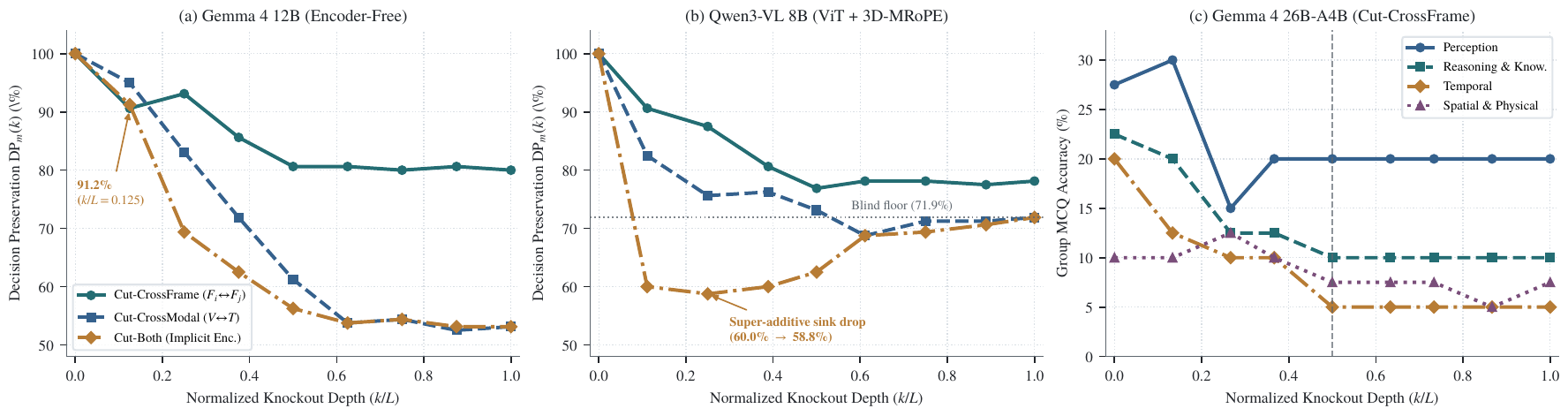}
\caption{\textbf{Intervention-mode contrast and capability-group breakdown.} \textbf{(a)} In encoder-free Gemma~4~12B, \cutboth preserves $91.2\%$ of baseline decisions at $k/L{=}0.125$, matching single-axis knockouts (\cutcf $90.6\%$, \cutcm $95.0\%$). \textbf{(b)} In ViT-equipped Qwen3-VL~8B, blocking cross-modal and cross-frame attention together causes a super-additive drop to $60.0\%$ ($k/L{=}0.11$) and $58.8\%$ ($k/L{=}0.25$), worse than cutting all $L$ layers ($71.9\%$). \textbf{(c)} Across \vidindex capability groups on Gemma~4~26B-A4B, \cutcf lowers \emph{Temporal} accuracy from $20.0\%$ to $5.0\%$ ($-75\%$ relative) and \emph{Reasoning \& Knowledge} from $22.5\%$ to $10.0\%$ ($-55.6\%$) by $k/L{=}0.50$, and \emph{Perception} retains $20.0\%$ of its $27.5\%$ baseline.}
\label{fig:modes_and_categories}
\end{figure*}

\subsection{Joint Isolation: Encoder-Free versus ViT-Equipped Models}
\label{sec:exp_both}

\Cref{fig:main_results}b and \Cref{fig:modes_and_categories}a--b compare \cutboth against its two single-axis components. This comparison cleanly separates the encoder-free model from the ViT-equipped models.

In encoder-free Gemma~4~12B-it, blocking both axes up to $k = 6$, or $k/L = 0.125$, preserves $91.2\%$ of decisions and $95.8\%$ of baseline accuracy, moving from $15.00\%$ to $14.37\%$. This closely matches single-axis masking at $90.6\%$ under \cutcf and $95.0\%$ under \cutcm. Beyond $k = 6$, \cutboth falls to $69.4\%$ at $k = 12$ and $53.1\%$ at $k = 48$. The first eighth of the decoder, $K_{85}^{\text{both}} = 6$ layers, therefore encodes each frame independently. Gemma~4~E4B-it likewise keeps $85.0\%$ at $k = 5$, and 26B-A4B-it keeps $80.0\%$ at $k = 4$.

ViT-equipped Qwen3-VL behaves differently: joint masking triggers a \emph{super-additive drop} in \Cref{fig:modes_and_categories}b. On Qwen3-VL~8B-Instruct at $k = 4$, or $k/L = 0.11$, single-axis knockouts preserve $90.6\%$ of decisions under \cutcf and $82.5\%$ under \cutcm, with full accuracy retention. Yet \cutboth preserves only $60.0\%$ at $k = 4$ and $58.8\%$ at $k = 9$. Both values sit more than $12$ points below the $71.9\%$ blind floor obtained when vision is blocked across all $36$ layers. Qwen3-VL~30B-A3B shows this asymmetry at $k = 6$: \cutcf retains $89.4\%$ and \cutcm retains $86.2\%$, whereas \cutboth drops to $70.0\%$. Its cause remains undetermined.

\subsection{Temporal Drift Follows Attention Topology}
\label{sec:exp_mftd}

\Cref{fig:main_results}c plots \mftd across layers on $120$ videos. At the input embeddings, layer~$0$, $\mathrm{MFTD}(0)$ equals $0.0000$ in all five models, confirming exact frame alignment. Above layer~$0$, the two model families diverge.
\begin{itemize}
    \item \textbf{Full-attention Qwen3-VL.} On Qwen3-VL~8B, \mftd climbs from $0.0001$ at layer~$1$ to a single peak across layers $17$--$19$, reaching $0.0787$ at layer~$17$, before falling to $0.0110$ at layer~$35$. Qwen3-VL~30B-A3B follows the same shape slightly deeper, peaking across layers $25$--$30$ with a maximum of $0.0315$ at layer~$28$. In both models, peak drift coincides with the half-depth saturation of $\mathrm{DP}_{\text{cf}}$.
    \item \textbf{Hybrid sliding-window Gemma~4.} Gemma~4 interleaves five local sliding-window layers with each global layer~\citep{gemma2026gemma4}. Sixteen frames yield thousands of visual tokens, exceeding a local window span so cross-frame pairs interact mainly in global layers. Accordingly, \mftd rises in discrete steps. On Gemma~4~12B, drift jumps tenfold at layer~$12$, from $0.0015$ to $0.0150$, before rising again at layers $32$ and $47$--$48$. Gemma~4~E4B and 26B-A4B show similar global-layer peaks at layers $24$ and $19$, detailed in \Cref{tab:mftd_appendix}.
\end{itemize}
Reversing the frames lowers accuracy on four of the five models: $\Delta_{\text{rev}}$ reaches $+5.83\%$ on Gemma~4~26B-A4B, $+4.17\%$ on Gemma~4~12B and Qwen3-VL~8B, and $+2.50\%$ on Qwen3-VL~30B-A3B. These models are sensitive to temporal order on \vidindex. The weakest model, Gemma~4~E4B, scores $7.50\%$ forward versus $10.83\%$ reversed, for reasons not yet determined.

\begin{table}[t]
\centering
\caption{\textbf{\cutcf accuracy (\%) across \vidindex capability groups ($N{=}40$ per group, $T{=}16$ frames).} Each cell reports $\mathrm{Base} \to \mathrm{CF}@L$ accuracy. By $k/L{=}0.50$, temporal and reasoning accuracy reach their lowest values, and perception retains most of its baseline.}
\label{tab:capability_groups}
\footnotesize
\setlength{\tabcolsep}{3.5pt}
\begin{tabular}{@{}lccc@{}}
\toprule
\textbf{Group} & \textbf{Gemma 4} & \textbf{Gemma 4} & \textbf{Qwen3-VL} \\
 & \textbf{26B-A4B} & \textbf{12B-it} & \textbf{30B-A3B} \\
\midrule
Perception & 27.5 $\to$ 20.0 & 20.0 $\to$ 17.5 & 20.0 $\to$ 12.5 \\
Reasoning & 22.5 $\to$ 10.0 & 10.0 $\to$ 10.0 & 10.0 $\to$ 7.5 \\
Spatial/Phys. & 10.0 $\to$ 7.5 & 20.0 $\to$ 7.5 & 5.0 $\to$ 0.0 \\
Temporal & 20.0 $\to$ 5.0 & 10.0 $\to$ 5.0 & 7.5 $\to$ 5.0 \\
\midrule
\textbf{Macro Avg.} & 20.0 $\to$ 10.6 & 15.0 $\to$ 10.0 & 10.6 $\to$ 6.2 \\
\bottomrule
\end{tabular}
\end{table}

\subsection{Capability Groups and Benchmark Audit}
\label{sec:exp_cats}

\Cref{tab:capability_groups} and \Cref{fig:modes_and_categories}c show that cross-frame knockout affects capability groups unevenly. On our strongest model, Gemma~4~26B-A4B-it at $20.00\%$ overall accuracy, blocking cross-frame attention up to $k/L = 0.50$ lowers \emph{Temporal} accuracy from $20.0\%$ to $5.0\%$, a $75\%$ relative drop, and lowers \emph{Reasoning \& Knowledge} from $22.5\%$ to $10.0\%$. Both groups then stay flat through layer~$L$. By contrast, \emph{Perception} retains $20.0\%$ of its $27.5\%$ baseline, as expected for appearance questions. Gemma~4~12B-it shows the same split: \emph{Spatial \& Physical} drops from $20.0\%$ to $7.5\%$ and \emph{Temporal} from $10.0\%$ to $5.0\%$, while \emph{Perception} shifts only from $20.0\%$ to $17.5\%$.

\paragraph{Low baseline accuracy.}
Zero-CoT accuracy spans $7.50\%$ to $20.00\%$. We identify two contributing factors.
\begin{enumerate}
    \item \textbf{Adversarial item selection.} \vidindex~\citep{song2026videoindex} uses Qwen3-VL~8B in its attack pyramid to purge items solvable from short context or a single frame. On the remaining items, single-pass 16-frame models often converge on the same wrong choice. Across $84$ of the $160$ items, at least four of our five models pick the same option, yet only $7$ of those $84$ consensus choices, or $8.3\%$, are correct.
    \item \textbf{Option-letter bias in the prompt.} Questions in this subset have up to eight options, and $35$ of the $160$ answers, or $21.9\%$, fall in E--H. Because our prompt ended with the example ``(e.g., A, B, C, or D)'', $93.1\%$--$96.2\%$ of predictions fell in A--D. Accuracy was $0.00\%$--$5.71\%$ on E--H items versus $9.60\%$--$24.00\%$ on A--D items, reaching $24.00\%$ on Gemma~4~26B-A4B. Because $\mathrm{DP}_m(k)$ and \mftd compare a model with itself under one prompt, this bias affects both sides equally. Our released script lists each item's options.
\end{enumerate}

\begin{table}[t]
\centering
\caption{\textbf{Early in-fusion isolation depths and super-additive drop on \vidindex ($N{=}160$).} $K_{85}/L$ reports the deepest swept layer fraction preserving $\mathrm{DP}_m(K) \ge 85\%$ under \cutcf, \cutcm, and \cutboth. $\Delta_{\text{sink}}(k_1)$ measures the super-additive drop at the first cutoff ($k_1 \approx 0.12L$): Gemma~4 models show values near zero, and ViT-equipped Qwen3-VL models drop by 16--23 points.}
\label{tab:infusion_bounds}
\footnotesize
\setlength{\tabcolsep}{3.0pt}
\begin{tabular}{@{}lcccc@{}}
\toprule
\textbf{Model} & $K_{85}^{\text{cf}}/L$ & $K_{85}^{\text{cm}}/L$ & $K_{85}^{\text{both}}/L$ & $\Delta_{\text{sink}}(k_1)$ \\
\midrule
Gemma~4~E4B-it & 23.8\% & 0.0\% & 11.9\% & $-0.62\%$ \\
Gemma~4~12B-it & 37.5\% & 12.5\% & 12.5\% & $-0.62\%$ \\
Gemma~4~26B-A4B & 0.0\% & 13.3\% & 0.0\% & $+1.88\%$ \\
Qwen3-VL~8B & 25.0\% & 0.0\% & 0.0\% & $+22.50\%$ \\
Qwen3-VL~30B-A3B & 12.5\% & 12.5\% & 0.0\% & $+16.25\%$ \\
\bottomrule
\end{tabular}
\end{table}

\subsection{Isolation Depths and Super-Additive Drop}
\label{sec:exp_cache}

\Cref{tab:infusion_bounds} summarizes the isolation depths $K_{85}/L$ and super-additive drops $\Delta_{\text{sink}}(k_1)$ defined in \Cref{eq:dsink}. Under \cutcf, isolation extends to $37.5\%$ of layers on Gemma~4~12B-it, $25.0\%$ on Qwen3-VL~8B, $23.8\%$ on Gemma~4~E4B, and $12.5\%$ on Qwen3-VL~30B-A3B. Under \cutboth, only the dense Gemma~4 models maintain a non-zero prefix: $12.5\%$ of layers, or $6/48$, on 12B and $11.9\%$, or $5/42$, on E4B. For these two models, frame states over the first eighth of layers can be cached independently of the prompt. Finally, the super-additive drop cleanly separates the two families. It remains near zero for Gemma~4, at $-0.62\%$ on E4B and 12B and $+1.88\%$ on 26B-A4B, whereas Qwen3-VL suffers large drops of $+22.50\%$ on 8B and $+16.25\%$ on 30B-A3B.

\section{Discussion}
\label{sec:discussion}

\paragraph{Lessons for native video early in-fusion.}
On static images, compute-matched pretraining shows that dedicated vision capacity in the early decoder layers matches a standalone encoder and outperforms full-depth modality separation~\citep{brown2026unbundling}. Our measurements show that existing video decoders already place frame encoding and cross-frame binding in their lower layers. We draw three lessons for training native video models.
\begin{enumerate}
    \item \textbf{Limit dedicated visual capacity to the lower half.} In all five models, blocking cross-frame attention beyond $k/L \approx 0.50$ alters few decisions, and accuracy losses on \emph{Temporal} and \emph{Reasoning} items occur below the midpoint (\Cref{fig:main_results}a, \Cref{fig:modes_and_categories}c). Native video models, whether using dense early in-fusion blocks or sparse MoE visual experts, can therefore omit modality-specific capacity above $0.5L$.
    \item \textbf{Verify joint visual isolation before relying on it.} Encoder-free Gemma~4~12B keeps $91.2\%$ of its decisions under \cutboth up to $k/L = 0.125$, with a super-additive drop of $-0.62\%$. Its shallow layers encode each frame independently of other frames and of the prompt, permitting prompt-agnostic frame caching. By contrast, Qwen3-VL loses $16.25$--$22.50$ points under \cutboth in \Cref{tab:infusion_bounds}, for reasons not yet determined. Any design that isolates early visual tokens should therefore check $\mathrm{DP}_{\text{both}}$ against the full-blind baseline. Attention-sink~\citep{xiao2024streamingllm} or register tokens~\citep{darcet2024registers} and separate visual pre-norms are candidate remedies.
    \item \textbf{Place global attention layers in the lower half of hybrid decoders.} In full-attention decoders, \mftd peaks across $\ell/L \in [0.47, 0.58]$. In 5:1 sliding-window decoders~\citep{beltagy2020longformer,gemma2026gemma4}, drift instead jumps at individual global layers in \Cref{fig:main_results}c. A native video model built on a sliding-window backbone should therefore place enough global layers below $0.5L$ so cross-frame binding finishes in the lower half.
\end{enumerate}

\paragraph{Limitations and Methodological Boundaries.}
Uniform sampling of $16$ frames and single-step logit scoring cannot capture dense micro-actions or multi-step reasoning. Our lessons also come from trained models evaluated at inference time. Validating them in end-to-end native video pretraining, and measuring wall-clock caching gains in vLLM~\citep{kwon2023vllm} or SGLang~\citep{zheng2024sglang}, remain future work.

\section{Conclusion}
\label{sec:conclusion}

We have shown that existing video MLLMs are implicitly early in-fusion. Across five models spanning dense, sparse MoE, ViT-equipped, and encoder-free designs, cross-frame attention affects decisions only in the lower half of the network. In its first eighth of layers, encoder-free Gemma~4~12B already encodes each frame independently of other frames and of the prompt. Yet the super-additive drop in Qwen3-VL shows that joint visual isolation is not automatically safe, and sliding-window decoders concentrate temporal drift at global layers. These findings guide three choices for native video training: how deep dedicated visual capacity should extend, when early visual tokens can be isolated, and where global attention layers belong.

\section*{Contributors}
Luoyuan Zhang, Xuanyu Zheng, Yancheng Long, Haoyang Xu, Bin Wen, Tingting Gao

\FloatBarrier
{\setlength{\bibsep}{3.5pt plus 0.5pt}\interlinepenalty=10000
\bibliographystyle{plainnat}
\bibliography{main}}

\clearpage
\appendix
\setcounter{table}{0}
\setcounter{figure}{0}
\renewcommand{\thetable}{A\arabic{table}}
\renewcommand{\thefigure}{A\arabic{figure}}
\renewcommand{\theHtable}{appendix.\arabic{table}}
\renewcommand{\theHfigure}{appendix.\arabic{figure}}
\section{Implementation Details}
\label{app:impl}

\subsection{Preserving Native Attention Masks}
Video MLLMs impose distinct constraints in their native attention masks $M_{\text{native}}^{(\ell)}$. Gemma~4 alternates five local sliding-window layers with one global layer and uses bidirectional attention within each image~\citep{gemma2026gemma4}. Qwen3-VL applies interleaved 3D MRoPE to packed visual tokens~\citep{bai2025qwen3vl}. Our \texttt{AttentionKnockoutContext} registers forward pre-hooks on layers $\ell < k$ and writes disallowed entries directly into the incoming 4D mask tensor. Sliding windows and intra-frame bidirectional attention are preserved, and $D_{i,i} = 0$ holds throughout. For Gemma~4~E4B-it, whose audio tower lacks PyTorch SDPA support in \texttt{transformers}, the loader selects SDPA or eager per submodule.

\subsection{Leakage in Ungrouped Linear Probing}
In our initial run, each video $n$ contributed two samples to the probe: $(\bar{v}_{\text{fwd}, n}^{(\ell)}, y=1)$ and $(\bar{v}_{\text{rev}, n}^{(\ell)}, y=0)$. Splitting these samples with \texttt{StratifiedKFold} failed to group by video. At layer~$0$, frame mean pooling makes the two vectors identical up to floating-point rounding, so $\bar{v}_{\text{fwd}, n}^{(0)} = \bar{v}_{\text{rev}, n}^{(0)}$. Whenever a pair splits across train and validation folds, the classifier assigned the validation twin its training label, which is always inverted. Layer-0 accuracy therefore fell to $7.5\%$ against a $50\%$ chance baseline, and peak validation accuracy stayed within $20.4\%$--$30.0\%$. Our released notebook groups folds by video via \texttt{GroupKFold(n\_splits=5)} and probes the paired difference $\bar{v}_{\text{fwd}, n}^{(\ell)} - \bar{v}_{\text{rev}, n}^{(\ell)}$. This paper reports only \mftd, $\mathrm{DP}_m(k)$, and the reversal drop, unaffected by this leakage.

\subsection{Layer-Wise \mftd{} Peaks}
\Cref{tab:mftd_appendix} compares each model's largest interior \mftd at $\ell < L$ against its final-layer value at $\ell = L$. Gemma~4~12B-it shows an early global-layer pulse of $0.0150$ at layer~$12$, up from $0.0015$ at layer~$11$ and $0.0022$ at layer~$13$, before reaching its largest values of $0.0297$--$0.0308$ at layers $47$--$48$. Both Qwen3-VL models peak near mid-depth, at layers $17/36$ and $28/48$, and decay toward layer $L{-}1$.

\begin{table}[h]
\centering
\caption{\textbf{Largest interior ($\ell < L$) and final-layer ($\ell = L$) \mftd{} of each model ($N{=}120$).}}
\label{tab:mftd_appendix}
\footnotesize
\setlength{\tabcolsep}{5pt}
\begin{tabular}{@{}lcccc@{}}
\toprule
\textbf{Model} & $L$ & \textbf{Int.\ Peak \mftd} & \textbf{Peak Layer ($\ell/L$)} & \textbf{Final $\ell{=}L$} \\
\midrule
Gemma~4~E4B-it & 42 & 0.0574 & 24 (0.57) & 0.0786 \\
Gemma~4~12B-it & 48 & 0.0297 & 47 (0.98) & 0.0308 \\
Gemma~4~26B-A4B-it & 30 & 0.0587 & 19 (0.63) & 0.0209 \\
Qwen3-VL~8B & 36 & 0.0787 & 17 (0.47) & 0.0528 \\
Qwen3-VL~30B-A3B & 48 & 0.0315 & 28 (0.58) & 0.0278 \\
\bottomrule
\end{tabular}
\end{table}

\end{document}